\documentclass[conference]{IEEEtran}
\IEEEoverridecommandlockouts
\usepackage{tabularray}
\usepackage{cite}

\def\bng{\bngx}

\font\bngx=bang10

\def\*#1*#2{o\null{#2}{#1}}

\def\sh#1{\setbox0=\hbox{#1}%
     \kern-.02em\copy0\kern-\wd0
     \kern.04em\copy0\kern-\wd0
     \kern-.02em\raise.0433em\box0 }
\usepackage[T1]{fontenc}
\usepackage{amsmath,amssymb,amsfonts}
\usepackage{algorithmic}
\usepackage{float}
\usepackage{graphicx}
\usepackage{textcomp}
\usepackage{xcolor}
\usepackage{hyperref}
\usepackage{multirow}
\usepackage{soul}
\usepackage{subcaption}
\usepackage{float}
\usepackage{booktabs,chemformula}
\usepackage{hyperref}
\usepackage{caption}
\usepackage{cleveref}
\usepackage{fancyhdr}
\usepackage[super]{nth}
\usepackage[pscoord]{eso-pic}
\usepackage{amssymb}

\def\BibTeX{{\rm B\kern-.05em{\sc i\kern-.025em b}\kern-.08em
    T\kern-.1667em\lower.7ex\hbox{E}\kern-.125emX}}

\newcommand\copyrighttext{%
  \footnotesize \textcopyright 2023 IEEE. Personal use of this material is permitted.
  Permission from IEEE must be obtained for all other uses, in any current or future
  media, including reprinting/republishing this material for advertising or promotional
  purposes, creating new collective works, for resale or redistribution to servers or lists, or reuse of any copyrighted component of this work in other works.}
\newcommand\copyrightnotice{%
\begin{tikzpicture}[remember picture,overlay]
\node[anchor=south,yshift=10pt] at (current page.south) {\fbox{\parbox{\dimexpr\textwidth-\fboxsep-\fboxrule\relax}{\copyrighttext}}};
\end{tikzpicture}%
}

\makeatletter

\let\old@ps@IEEEtitlepagestyle\ps@IEEEtitlepagestyle
\def\confheader#1{%
    \def\ps@IEEEtitlepagestyle{%
        \old@ps@IEEEtitlepagestyle%
        \def\@oddhead{\strut\hfill#1\hfill\strut}%
        \def\@evenhead{\strut\hfill#1\hfill\strut}%
    }%
    \ps@headings%
}
\makeatother

\confheader{%
    \parbox{50cm}{\textbf{Accepted and presented at "2023 26th International Conference on\\ Computer and Information Technology (ICCIT) 13-15 December 2023, Cox’s Bazar, Bangladesh"}}
}

\begin{document}

\title{Bengali Intent Classification with Generative Adversarial BERT  \\
}


 \author{
    \IEEEauthorblockN{Mehedi Hasan\IEEEauthorrefmark{1}\textsuperscript{\textsection}, 
     Mohammad Jahid Ibna Basher\IEEEauthorrefmark{2}\textsuperscript{\textsection}, 
     and Md. Tanvir Rouf Shawon\IEEEauthorrefmark{1}\textsuperscript{\textsection}
  }
 \IEEEauthorblockA{\IEEEauthorrefmark{1} Department of Computer Science and Engineering, Ahsanullah University of Science and Technology \\}
 \IEEEauthorblockA{\IEEEauthorrefmark{2} Department of Computer Science and Engineering, Chittagong University of Engineering and Technology\\}
abir.aust.102@gmail.com,
  jahidibnabasher@gmail.com,
  shawontanvir95@gmail.com}

\maketitle
\copyrightnotice
    \begingroup\renewcommand\thefootnote{\textsection}
        \footnotetext{All authors contributed equally to this work.}
    \endgroup
\begin{abstract}
Intent classification is a fundamental task in natural language understanding, aiming to categorize user queries or sentences into predefined classes to understand user intent. The most challenging aspect of this particular task lies in effectively incorporating all possible classes of intent into a dataset while ensuring adequate linguistic variation. Plenty of research has been conducted in the related domains in rich-resource languages like English. 
In this study, we introduce BNIntent30, a comprehensive Bengali intent classification dataset containing 30 intent classes. The dataset is excerpted and translated from the CLINIC150 dataset containing a diverse range of user intents categorized over 150 classes. Furthermore, we propose a novel approach for Bengali intent classification using Generative Adversarial BERT to evaluate the proposed dataset, which we call GAN-BnBERT. Our approach leverages the power of BERT-based contextual embeddings to capture salient linguistic features and contextual information from the text data, while the generative adversarial network (GAN) component complements the model's ability to learn diverse representations of existing intent classes through generative modeling. Our experimental results demonstrate that the GAN-BnBERT model achieves superior performance on the newly introduced BNIntent30 dataset, surpassing the existing Bi-LSTM and the stand-alone BERT-based classification model.


\end{abstract}


\section{Introduction}
Chatbots, voice assistants, and contact centers all use autonomous question-answering systems to
minimize the amount of time a human has to respond. The advancement in this field of study in the last few years has been remarkable. Numerous studies \cite{kanodia2021question,lovato2019hey} on autonomous Q/A systems have been conducted and are currently being used in real-time by Google, Apple, Microsoft, and other major corporations. The Q/A system must detect the user's intent or purpose precisely to provide answers that are more relevant to the user. Research in this area has been conducted in a variety of languages, with the majority of the studies being conducted in English.
\cite{larson2019evaluation,chen2019bert,zhou2022knn}.

The Q/A systems are not well-trained to perform appropriately in languages with limited resources due to a lack of proper understanding of the user's intent. There is no noteworthy study for intent categorization in Bengali. The lack of benchmark datasets, resources, and language processing tools has led to a significant gap in research within the domain of Bengali intent classification. The primary objective of
\begin{figure}[!hbtp]
  \centering
  \includegraphics[scale=1.2]{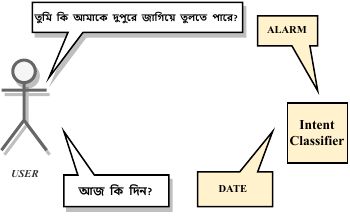}
  \caption{Sample user utterances (white box) and the model's determination of the various intent classes (yellow box). The \textit{top} sample is categorized as being in the ALARM class, whereas the \textit{bottom} sample is in the DATE class.}
  \label{fig:abstract}
\end{figure}
this research aims to provide a dataset for Bengali intent classification that can be a massive aid to conversational AI systems. Following our goal, we created a dataset that was translated from a well-known English dataset named CLINC150 \cite{larson2019evaluation}. We call our dataset \textbf{BNIntent30}. 

To enhance the distribution of semantic meaning across all classes in our dataset and obtain a benchmark result for our intent classification task, the Bidirectional Encoder Representations from Transformers (BERT) language model has been effectively integrated with a semi-supervised GAN, commonly known as GAN-BERT \cite{Croce2020}. Fig. \ref{fig:abstract} depicts a graphical representation of our work. The contributions of this work can be summarized in the following points.
\begin{itemize}
    \item We have proposed a Bengali intent dataset by translating and curating a well-known English intent dataset called CLINC150 \cite{larson2019evaluation}  with 30 different classes from the 150 classes of CLINC150. We name our dataset \textbf{\textit{BNIntent30}}.  
    \item On top of the BERT model, we have proposed a semi-supervised GAN architecture called \textbf{\textit{GAN-BnBERT}} to categorize the user's intent when interacting with an autonomous question-answering agent.
    \item A conventional architecture, BiLSTM, and a stand-alone BERT model were also studied. The models were tested on many test samples, with the best-performing model being inferred. The implementation and dataset can be found at - \url{https://www.github.com/Jahid006/GAN-BnBERT}
\end{itemize}

\section{Related Work}
This section uses two different sub-sections to highlight the literature review on intent classification and the usage of generative adversarial network-driven language model 
in the downstream NLP tasks.

\subsection{Study on Intent Classification}
The intent of a user when speaking with a conversational agent has been categorized in several works. In addition to 150 intent classifications spread over 10 distinct domains, Stefan Larson et al. \cite{larson2019evaluation} developed their dataset to classify out-of-scope intent. To assess their dataset, the author also gave several benchmark models. Bidirectional Encoder Representation of Transformers (BERT), one of the benchmarking models, achieves the maximum accuracy of 96.7\% for 150 classes that are within its scope and the out-of-scope class. 

An intent classification model with joint slot filling was demonstrated in the work by Qian Chen \cite{chen2019bert}. They have conducted experiments using many publicly accessible datasets and have had successful outcomes. Using the Snips and ATIS datasets, the joint BERT model classified intent with an accuracy of 98.6\% and 97.5\%, respectively.

A contrastive learning-based strategy using a combination of KNN classifiers \cite{zhou2022knn} has been presented by Yunhua Zhou et al. to identify out-of-distribution intent. With the CLINC-FULL, CLINC-SMALL, BANKING, and StackOverflow datasets, they have produced striking results. In-domain intent classification results for the CLINC-FULL and CLINC-SMALL datasets were 92.61\% and 91.92\%, respectively.

Two distinct approaches for multiclass intent classification were proposed by Hemant Purohit et al. \cite{7463729} and Jetze Schuurmans et al. \cite{schuurmans2019intent}, who also achieved noteworthy results using multiple data sets. 

The study \cite{firdaus2023multitask} by Mauajama Firdaus et al. focused on multilingual intent classification, whereas the study \cite{9415112} by Akshat Gupta et al. focused on acoustic-based intent classification for low-resource languages such as Hindi and Bengali.

\subsection{Study on Generative Adversarial BERT}

In their study \cite{Croce2020}, Danilo Croce et al., who used generative adversarial networks on top of BERT (LM) for classification tasks, initially suggested GAN-BERT. The concept of semi-supervised learning was expanded upon in this study by the researchers in the context of classification. They demonstrated the effectiveness of their methodology in many downstream tasks, including sentiment analysis, news categorization, and question classification.

In their research \cite{ta2022gan}, Hoang Thang Ta et al. classified aggressive and violent incidents using the GAN-BERT model. They obtained an acceptable F1-Score of F1 of 74.43\% using tweets gathered from social media platforms.

There have been studies done in Bengali using the GAN-BERT architecture. Raihan Tanvir et al. \cite{tanvir2022gan} used it to detect fake news and hate speech in Bengali, achieving 75.4\% and 92.6\% accuracy on the task while addressing the issue of a small number of data in the datasets. In the article \cite{10236810} by Shawon et al. used GAN-LM architecture and their dataset to classify fake reviews in Bengali, achieving an accuracy of 83.59\% with a limited amount of data for each class.

\section{Background Study} 
The following sections delve into the investigated architectures along with the metrics that are used to evaluate the performance of these architectures.

\textbf{Bidirectional Long Short Term Memory (BiLSTM):} BiLSTM is proposed as an expansion of Long short-term memory architecture in both the left-to-right and right-to-left directions for grasping the context of words from both sides. The LSTM architecture solves the vanishing gradient issue in recurrent neural networks, making it a reliable architecture to start with.

\textbf{Bidirectional Encoder Representation of Transformers (BERT):} BERT \cite{devlin2018bert} is the first fine-tuned model that can be applied to a variety of downstream tasks, including sentiment analysis, categorization, and question-answering. With an attention mechanism \cite{vaswani2017attention}, it simulates the Masked Language Model (MLM). During the training procedure, 15\% of the tokens are randomly masked to help the model become accustomed to the similar embeddings of a single token. It is more practical for many NLP tasks thanks to the concept of left-to-right and right-to-left training of language models.

\textbf{Semi-supervised Generative Adversarial Network:} A fairly well-known architecture in the field of generative models is the generative adversarial network (GAN) \cite{goodfellow2020generative}. A solitary GAN model is enhanced by a semi-supervised GAN \cite{salimans2016improved} model. It is a strategy to train the generator to produce more realistic data by giving it a Gaussian distribution with the shape $N\left(\mu, \sigma^2\right)$ to better prepare the discriminator for the detection task. The supervised data is given to the discriminator to train it to properly predict the real examples.

\textbf{Performance Metrics:} Our models are evaluated using the following performance metrics: accuracy, precision, recall, f1-score, and MCC score. The accuracy is defined as the ratio of correctly categorized samples among all testing samples. The precision score determines the ratio of all the correctly categorized samples to all the samples that are predicted positive. Recall emphasizes false negative (FN) cases and assesses the proportion of correctly classified samples among all the positive cases. F1 represents the harmonic mean of precision and recall. Instead, the Matthews correlation coefficient (MCC) is a more dependable statistical measure that only generates a high score if the prediction is successful in all four categories (TP, TN, FP, FN) of the confusion matrix.

\section{Dataset}
The dataset we are using is translated from the CLINC150 \cite{larson2019evaluation} dataset, where 30 classes out of 150 were taken into consideration. We name the dataset \textbf{\textit{BNIntent30}}

\subsection{Dataset Accumulation}

Due to the lack of benchmark datasets in the relevant domain, we were prompted to develop our dataset. However, to save resources and manpower, we finally decided to utilize any industry-standard translation system to transform an existing dataset from English Language to Bengali Language. We selected The CLINC150 \cite{larson2019evaluation} dataset for translation as it offers intent from a wide range of classes for Task-oriented dialog systems. The dataset was created primarily for out-of-scope categorization, with the authors accumulating 150 intent classes within ten domains. 

\begin{table}[]
\caption{
Data distribution over 30 classes in terms of class domain}
\label{tab:dist}
\begin{tabular}{cccccc}
\hline
\textbf{Domain}& \textbf{Class}           & \textbf{Train} & \textbf{Test} & \textbf{\begin{tabular}[c]{@{}c@{}}Valid-\\ ation\end{tabular}} & \textbf{Total} \\ \hline
\multirow{6}{*}{HOME}      & CALENDER    & 100& 30          & 20& 150\\ \cline{2-6} 
 & WHAT SONG   & 100& 30          & 20& 150\\ \cline{2-6} 
 & PLAY MUSIC  & 99& 30          & 19& 148\\ \cline{2-6} 
 & NEXT SONG   & 98& 30          & 20& 148\\ \cline{2-6} 
 & TODO LIST   & 98& 27          & 20& 145\\ \cline{2-6} 
 & REMAINDER   & 100& 30          & 20& 150\\ \hline
\multirow{7}{*}{UTILITY}   & DATE        & 100& 30          & 20& 150\\ \cline{2-6} 
 & TIME        & 100& 30          & 20& 150\\ \cline{2-6} 
 & ALARM       & 96& 30          & 20& 146\\ \cline{2-6} 
 & SPELLING    & 100& 30          & 20& 150\\ \cline{2-6} 
 & MAKE CALL   & 90& 28          & 20& 138\\ \cline{2-6} 
 & CALCULATOR  & 99& 27          & 20& 146\\ \cline{2-6} 
 & WEATHER     & 100& 30          & 19& 149\\ \hline
\multirow{8}{*}{\begin{tabular}[c]{@{}l@{}}SMALL\\ TALK\end{tabular}}          & THANK YOU   & 100& 30          & 20& 150\\ \cline{2-6} 
 & GOODBYE     & 95& 30          & 20& 145\\ \cline{2-6} 
 & \begin{tabular}[c]{@{}c@{}}HOW OLD\\ ARE YOU\end{tabular}       & 100& 30          & 20& 150\\ \cline{2-6} 
 & TELL JOKE   & 100& 30          & 20& 150\\ \cline{2-6} 
 & FUN FACT    & 98& 30          & 19& 147\\ \cline{2-6} 
 & \begin{tabular}[c]{@{}c@{}}WHERE ARE\\ YOU FROM\end{tabular}    & 100& 30          & 20& 150\\ \cline{2-6} 
 & \begin{tabular}[c]{@{}c@{}}WHAT ARE \\ YOUR HOBIES\end{tabular} & 100& 30          & 20& 150\\ \cline{2-6} 
 & \begin{tabular}[c]{@{}c@{}}WHAT IS \\ YOUR NAME\end{tabular}    & 99& 30          & 20& 149\\ \hline
\multirow{5}{*}{META}      & \begin{tabular}[c]{@{}c@{}}CHANGE \\ USER NAME\end{tabular}     & 100& 30          & 20& 150\\ \cline{2-6} 
 & \begin{tabular}[c]{@{}c@{}}CHANGE \\ VOLUME\end{tabular}        & 94& 28          & 20& 142\\ \cline{2-6} 
 & NO          & 94& 29          & 20& 143\\ \cline{2-6} 
 & REPEAT      & 99& 30          & 20& 149\\ \cline{2-6} 
 & YES         & 97& 30          & 20& 147\\ \hline
\multirow{3}{*}{\begin{tabular}[c]{@{}l@{}}AUTO\\ AND \\ COMMUTE\end{tabular}} & \begin{tabular}[c]{@{}c@{}}CURRENT\\ LOCATION\end{tabular}      & 99& 29          & 18& 146\\ \cline{2-6} 
 & TRAFFIC     & 97& 28          & 20& 145\\ \cline{2-6} 
 & DISTANCE    & 100& 30          & 20& 150\\ \hline
TRAVEL        & TRANSLATE   & 100& 30          & 20& 150\\ \hline
\end{tabular}
\end{table}
\subsection{Class Selection}
Here we select only 30 classes out of 150 classes of CLINIC150 dataset. The main criteria behind selecting the classes are personal queries or where users ask some questions like "What is your name? and "What are your hobbies?" or basic commands like "Turn on the music" "Increase the volume" etc. The 30 classes we select span 6 domains, with 5 of those domains holding the majority of the categories. Table \ref{tab:dist} shows the selected classes along with their domain. 
\subsection{Data Translation}
\begin{figure}[!hbtp]
  \centering
  \includegraphics[scale=.56]{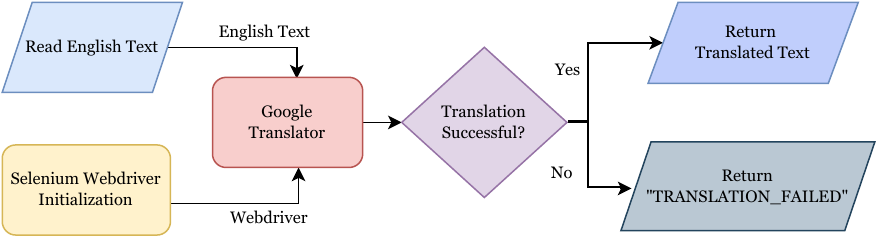}
  \caption{A schematic diagram of data translation procedure}
  \label{fig:trans}
\end{figure}

The dataset we utilize here consists of samples in English. To proceed, the chosen English data samples must be translated into Bengali for the selected classes. We used the Google Translator API to complete this operation, selecting Bengali as the target language and English as the source language. Fig. \ref{fig:trans} shows the data translation procedure graphically. After translating the data samples, we carefully reviewed every piece of data to weed out any discrepancies. We eliminated a few instances since their text lengths fell below 2. Table \ref{tab:translationsamples} shows some translated samples along with their English equivalent. 

\begin{table}[!h]
\centering
\caption{Some translation samples with their English equivalents.}
\label{tab:translationsamples}
\begin{tabular}{cc}
\hline
\textbf{English Text}     & \textbf{Bengali Text}           \\ \hline
\begin{tabular}[c]{@{}c@{}}will you help me with \\ a math problem\end{tabular}       & \begin{tabular}[c]{@{}c@{}}\bng Aapin ik Aamaek giNetr smsYa \\ \bng iney sHaJta krebn? \end{tabular} \\ \hline
\begin{tabular}[c]{@{}c@{}}what day is it gonna be in \\ twenty-one days\end{tabular} & \bng Ekush idenr medhY ekan idn Heb?
\\ \hline
\begin{tabular}[c]{@{}c@{}}i'm glad i got to talk \\ to you\end{tabular}              & \begin{tabular}[c]{@{}c@{}} \bng Aaim Aapnar saeth ktha blet \\ \bng eper Aanin/dt   \end{tabular}             \\ \hline
\begin{tabular}[c]{@{}c@{}}i didn't hear you please \\ repeat\end{tabular}            & \begin{tabular}[c]{@{}c@{}}\bng Aaim shuinin Aapin dya ker \\ \bng punrabrRit/t krun\end{tabular}        \\ \hline
\end{tabular}
\end{table}

\subsection{Dataset Statistics}
After translation and cleaning, there are 4433 total units of data, of which 2952 are in the training set, 595 are in the validation set, and the remaining 886 are in the testing set. Table \ref{tab:dist} shows the class-wise distribution of the data samples. Some important statistics of the converted dataset are shown in Table \ref{tab:stat}, where we can observe that the greatest length of each data is 128 and the minimum is only 2. Therefore, it is evident that the length of the data we are employing in our work is not very high. The length is 32.18 on average, which is adequate for any experiment.

\begin{table}[!h]
\centering
\caption{Statistics for the converted dataset of 30 classes}
\label{tab:stat}
\begin{tabular}{cc}
\hline
\textbf{Statistics} & \textbf{Number} \\ \hline
Total Words         & 25968           \\ \hline
Total Unique Words  & 3065\\ \hline
Maximum Length      & 128\\ \hline
Minimum length      & 2  \\ \hline
Average length      & 32.18           \\ \hline
\end{tabular}
\end{table}

\section{Methodology} \label{sec:methodology}
To classify the intent of a given text sequence, we followed GAN-BERT \cite{Croce2020} architecture that facilitates robust learning for a few labeled training examples, which incorporates a BERT-based classifier along with a GAN-based generator. The generator synthesizes intent-specific features, which are then combined with BERT embeddings to improve classification accuracy. We call our proposed architecture \textbf\textit{{GAN-BnBERT}}. The model is comprised of three main components:

\textbf{a) Language Model:} A BERT-based pretrained Language Model is utilized to extract embedding of the input sequence as pretrained LM are known to be excellent in understanding the textual semantics of any given text because of the pretraining strategies they employed such as Masked Language Modeling, Sentence Entrailment, etc. The LM we used projects the input text sequence to a fixed dimensional vector of size 768 which is later digested by the Discriminator to determine which class the text sequence belongs to.

\textbf{b) Generator: } The Generator block takes a Gaussian noise as input and generates an arbitrary representation of a fake class of the same shape as the Language Model. The generated vector could be classified as one of the real intent classes if the Discriminator thinks the representation is 'Closer' to that class.

\textbf{c) Discriminator: } This block is a Feed Forward Neural Network(FFN) that takes the real and fake embedding generated by the LM and the Generator respectively and classifies them into one of the real classes or the fake representation class.

All three modules are trained end-to-end during the training process while during the inference the Generator block is removed as we do not need to classify the fake representation class during the inference stage.

\begin{figure*}[!hbtp]
  \centering
  \includegraphics[scale=.75]{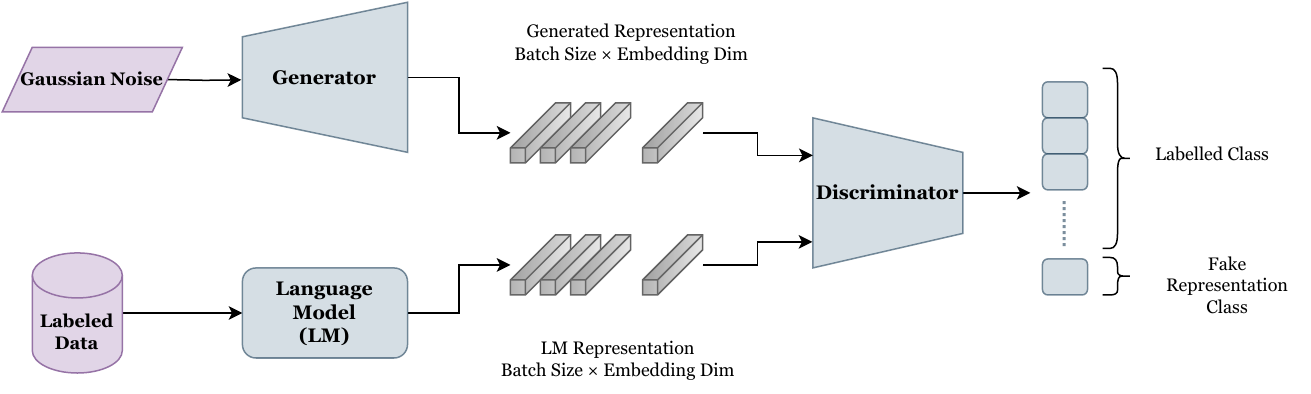}
  \caption{Illustrates the complete architecture of our proposed GAN-BnBERT architecture. The Discriminator is trained to classify the real(generated by LM from labeled data) and fake(generated by the Generator) representations into their respective classes.}
  \label{fig:methodology}
\end{figure*}

\section{Model Architecture}
The architecture of the models we employ is described in this section. BiLSTM contains a total of 6.80 million parameters with two BiLSTM layers and a single dense layer. A layer of embedding with an embedding dimension of 64 is the first layer in the model. There are 110 million trainable parameters in the BanglaBERT language model (LM) altogether \cite{bhattacharjee2021banglabert}. We fine-tuned the BanglaBERT model using our training data and validated it using a validation set. Section \ref{sec:methodology} describes the GAN-BnBERT architecture, which consists of a semi-supervised GAN model with a generator and a discriminator on top of the BanglaBERT language model.
\section{Experimental Result \& Analysis}
\begin{figure*}[htbp]
  \centering
  \subfloat{\includegraphics[width=2.3in]{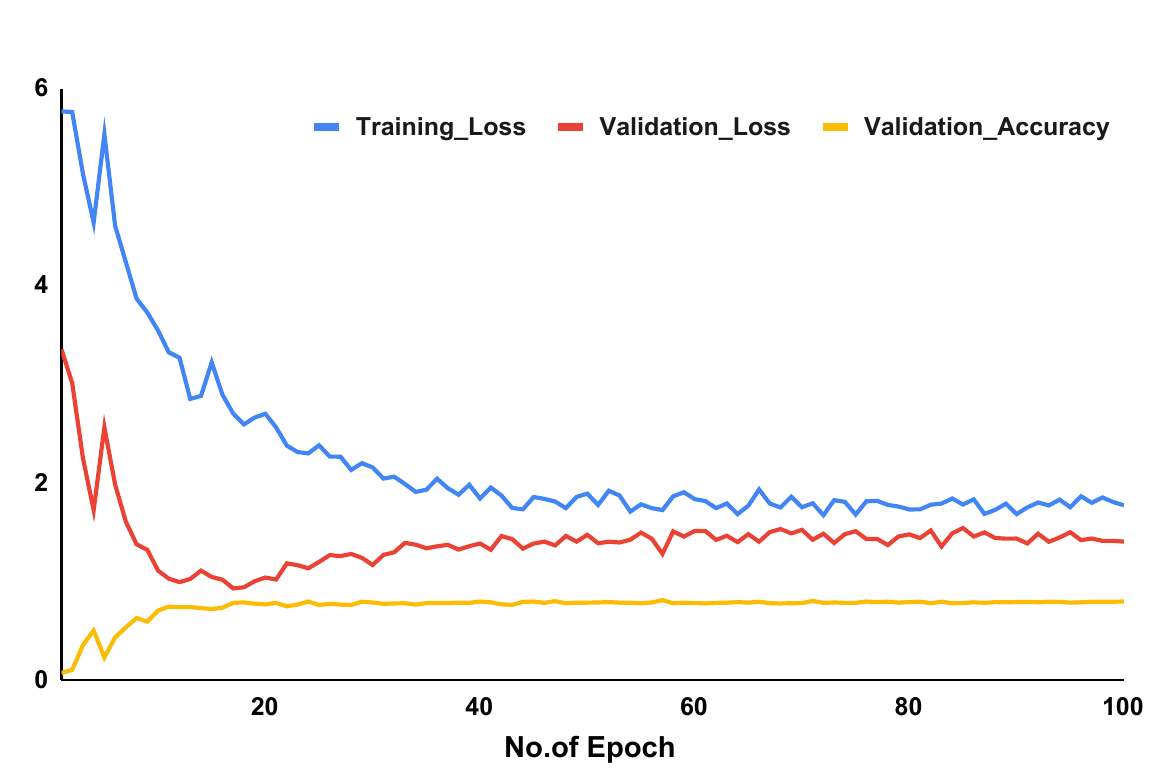}}\label{fig:image4}
  \subfloat{\includegraphics[width=2.3in]{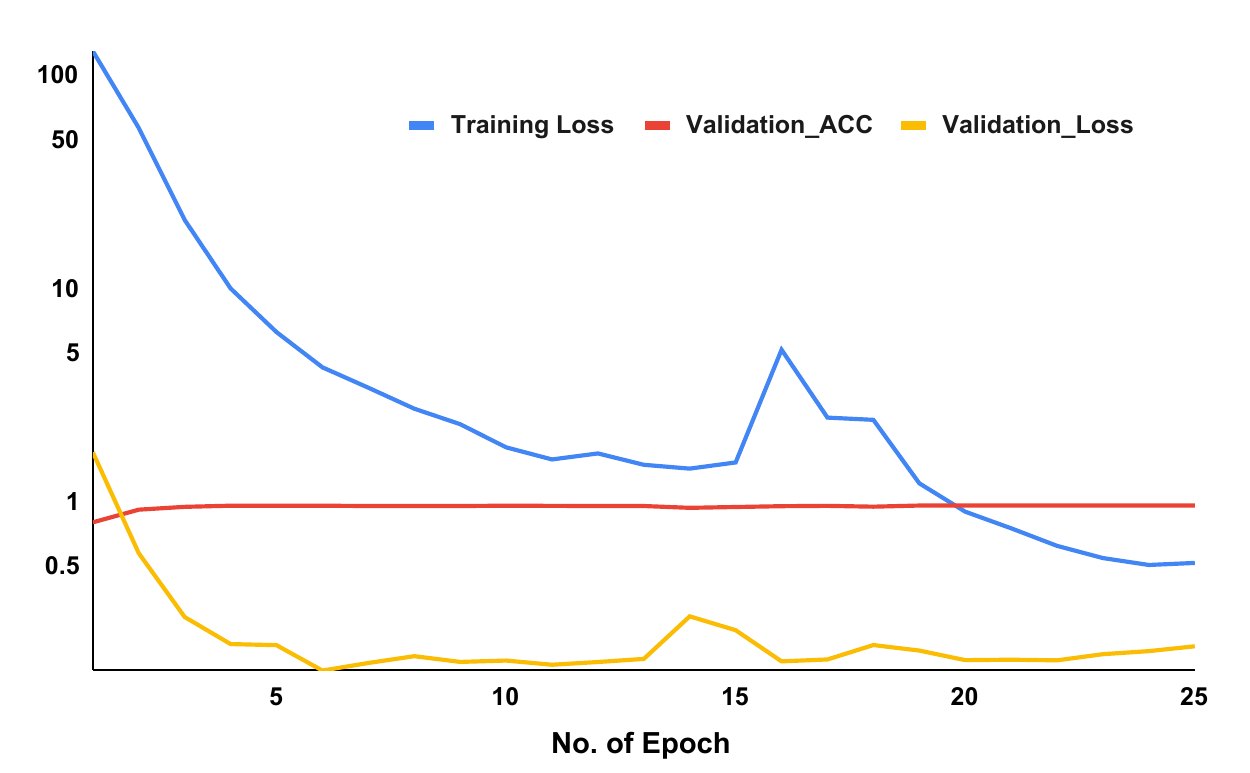}}\label{fig:image3}
  \subfloat{\includegraphics[width=2.3in]{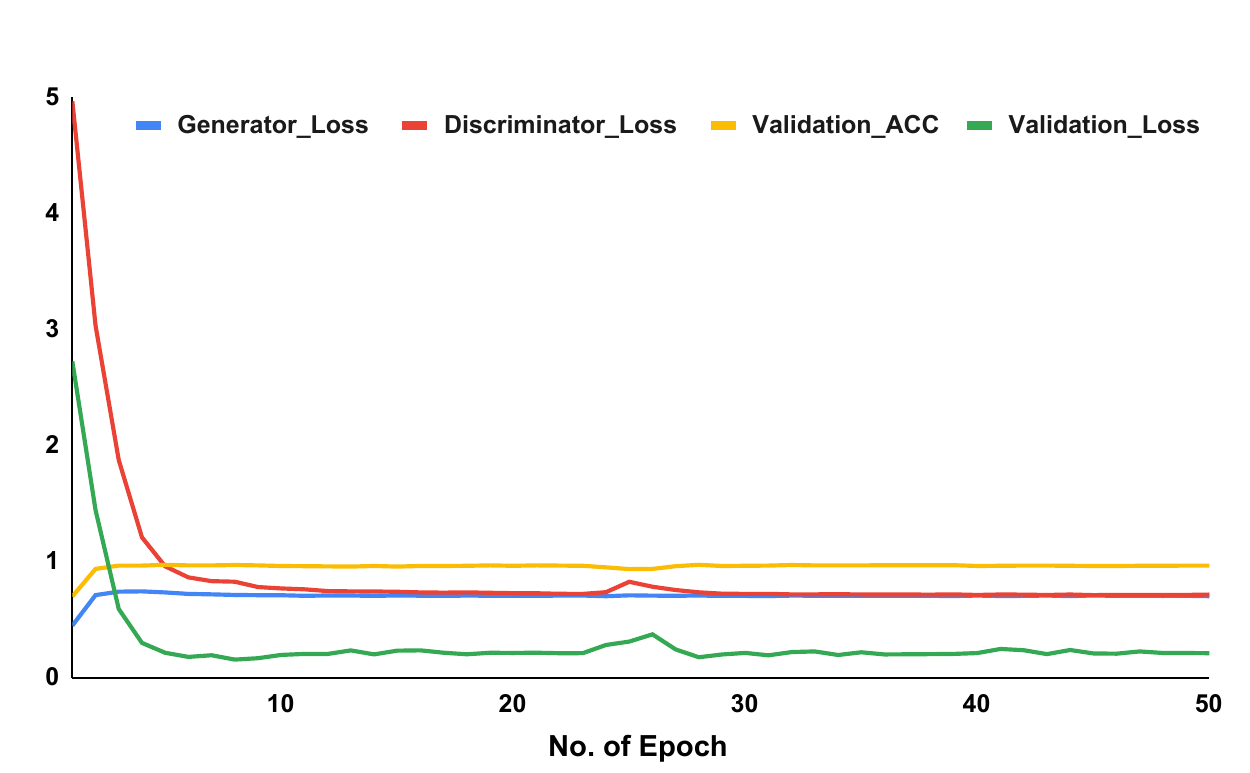}}\label{fig:image3}
 
  \caption{Loss vs accuracy graph of BiLSTM(\textit{left}), BanglaBERT(\textit{middle}) and GAN-BnBERT (\textit{right}).}
  \label{fig:loss}
\end{figure*}

\begin{figure}[]
  \centering

  \includegraphics[scale=.16]{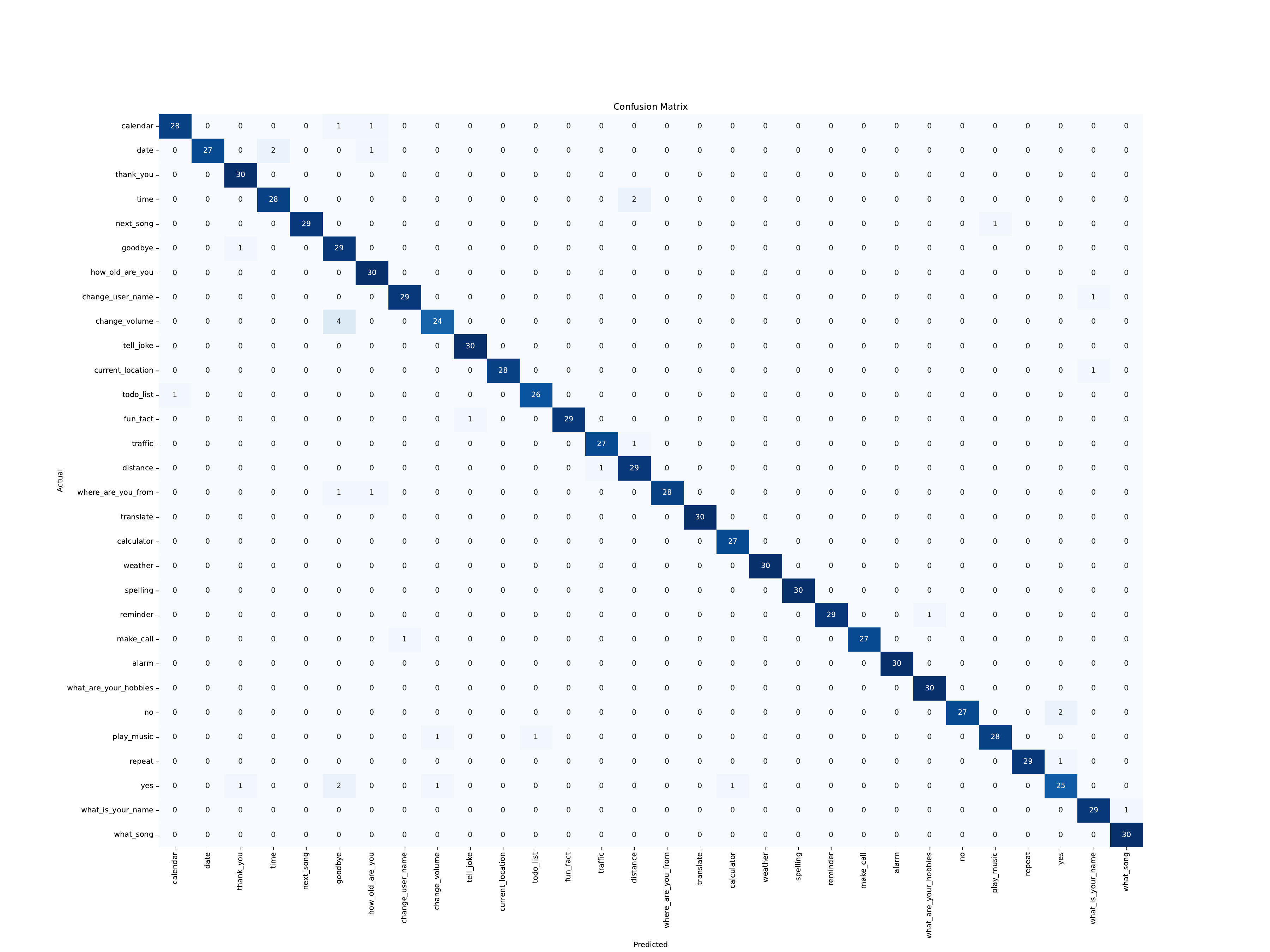}
    \caption{Confusion Matrix of GAN- BnBERT.}
  \label{fig:conf}
\end{figure}

The study and results of the examined model are presented in this section. The experimental results of numerous benchmark models utilizing different performance metrics are displayed in Table \ref{tab:result}. Fig. \ref{fig:loss} presents the loss vs accuracy graphs from the training period of BiLSTM, fine-tuned BERT, and GAN-BnBERT. The Confusion Matrix of the GAN-BnBERT model is illustrated in Fig. \ref{fig:conf}.  

\begin{table}[!h]
\caption{Experimental results of the explored models }
\label{tab:result}
\begin{tabular}{cccccc}
\hline
Model & \textbf{\begin{tabular}[c]{@{}c@{}}Accuracy \\ (\%)\end{tabular}} & \textbf{Precision} & \textbf{Reccall} & \textbf{F1} & \textbf{MCC}  \\ \hline
BiLSTM& 79.35 & 0.7979             & 0.7943           & 0.7915      & 0.7866          \\ \hline
BanglaBERT             & 96.05 & 0.9619             & 0.9604           & 0.9599      & 0.9592         \\ \hline
\begin{tabular}[c]{@{}c@{}}GAN-\\ BnBERT\end{tabular} & 96.73 & 0.9694             & 0.9672           & 0.9672      & 0.9662      \\ \hline
\end{tabular}
\end{table}

\subsection{Hyperparameter Settings}
The hyperparameters used in the explored models are listed in Table \ref{tab:hyper}. the optimal hyperparameters are discovered after systematically experimenting with a wide range of values. It is clear from the table that BiLSTM required 100 epochs to reach a saturation point, but GAN-BnBERT and BanglaBERT converged after 50 and 25, respectively. 
\begin{table}[!h]
\centering
\caption{Hyperparameter used in different model}
\label{tab:hyper}
\begin{tabular}{cccc}
\hline
\textbf{Architecture} & \textbf{BiLSTM} & \textbf{BanglaBERT}    & \textbf{\begin{tabular}[c]{@{}c@{}}GAN\\ BnBERT\end{tabular}} \\ \hline
Optimizer & Adam& AdamW& Adam        \\ 
Learning Rate         & 0..001          & 5e\textasciicircum{}-5 & 0.01        \\ 
Epoch     & 100 & 25   & 50          \\ 
Batch Size& 128 & 64   & 64          \\ 
Dropout Rate          & 0.2 & 0.1  & 0.2         \\ \hline
\end{tabular}
\end{table}


\subsection{Result Analysis}
\textbf{\underline{Quantitative Analysis:}}
Table \ref{tab:result} shows the performance of the models on various metrics, such as accuracy, precision, recall, F1 score, and MCC. It is clear from the table that the BiLSTM model exhibits a commendable accuracy of 79.35\% in intent classification. This hints at the possibility of improved performance outcomes from increasing learnable parameters. On the other hand, BanglaBERT, a prominent pre-trained model tailored for the Bengali language, demonstrates its effectiveness across diverse natural language processing tasks. It accomplishes a significantly improved accuracy rate of 96.05\% by harnessing its enriched pre-trained embeddings. Furthermore, GAN-BnBERT introduces a comprehensible refinement to the classification task, achieved through the augmentation process of Generative Adversarial Network (GAN). It is noteworthy that GAN-BnBERT achieves a remarkable 96.73\% accuracy, indicating its superiority over both BanglaLBERT and BILSTM counterparts for Bengali language intent classification. The findings suggest that the implementation of GAN stands to increase the effectiveness of the BERT models and cover a wider data spectrum. 

The confusion matrix in Fig. \ref{fig:conf} demonstrates that over 95\% of samples are correctly classified across all classes by GAN-BnBERT. The heightened diagonal values signify robust classification for each of the 30 intent classifications. Regardless, only a few misclassifications are evident, such as CHANGE\_VOLUME and YES classes, have more misclassifications than others. Interestingly, four samples from the CHANGE\_VOLUME class are assigned the GOOD\_BYE class. It could be due to the shared vocabulary between the two classes. The loss vs. accuracy graph for each of the three examined models is shown in Figure \ref{fig:loss}. The Bi-LSTM and fine-tuned BERT exhibit instability during the training period and converge after 100 and 25 epochs, respectively, while GAN-BnBERT exhibits stable training, as can be seen from the graphs.

\textbf{\underline{Qualitative Analysis:}}
To completely comprehend a model's behavior, quantitative investigation into the model is insufficient, and extensive analysis needs to be carried out to interpret the model's behavior with the test samples. To facilitate this, we provide a misclassification analysis using several test samples. Table \ref{tab:miss_sample} displays some samples that are incorrectly identified using our top-performing GAN-BnBERT model.  The target class in the first example is TIME, but the model forecasts it to be DISTANCE with a probability of 0.58, whereas TIME is the model's most likely nearby class with a chance of 0.40. It is clear that the model does not agree with the prediction in all cases. The term {\bng tanjainyay} could be confusing because it is a country name, but the word {\bng smy} forces the model to predict the class TIME. Similar circumstances apply to the second instance where the model predicts traffic with a relatively comparable probability despite the target class being DISTANCE. The terms {\bng Upay} and {\bng kachakaich} force the model to estimate TIME despite the user's request in this sentence for the distance to a location. 

The outcome of the third sample from table \ref{tab:miss_sample} is quite fascinating. Here, the intended class is DATE but is projected to be TIME with a probability value of 0.86. The next likely class is CURRENT LOCATION with a probability score of 0.11. The words {\bng Aaich} and {\bng ekan} may force the model to favor the TIME class and give a high score to the CURRENT LOCATION class, respectively. It is clear that the model is being greatly confused by a few common words from both the TIME and DATE classes.

\begin{table*}[]
\centering
\caption{Some instances of miss-classified samples along with the nearest probable class and probability}
\label{tab:miss_sample}
\begin{tabular}{cccccc}
\hline
\textbf{Test Sample} & \textbf{\begin{tabular}[c]{@{}c@{}}Target \\ Class\end{tabular}} & \textbf{\begin{tabular}[c]{@{}c@{}}Predicted \\ Class\end{tabular}} & \textbf{\begin{tabular}[c]{@{}c@{}}Probability of\\ Predicted Class\end{tabular}} & \textbf{\begin{tabular}[c]{@{}c@{}}Nearest \\ Probable \\Class\end{tabular}} & \textbf{\begin{tabular}[c]{@{}c@{}}Probability of \\ nearest probable \\ class\end{tabular}} \\ \hline
\begin{tabular}[c]{@{}c@{}}\bng E{I} muHuer/t Aamaek tanjainyay smy idn\\ (please give me the time in tanzania at this moment)\end{tabular}       & TIME & DISTANCE& 0.58 & TIME            & 0.40             \\ \hline
\begin{tabular}[c]{@{}c@{}}\bng Aamar ik Ja{O}yar Upay Aaech ba Aaim epn es/Tshen Ja{O}yar \\ \bng kachakaich Aaich\\ (do i have a ways to go or am i close to getting to penn station)\end{tabular} & DISTANCE      & TRAFFIC & 0.50 & DISTANCE        & 0.45             \\ \hline
\begin{tabular}[c]{@{}c@{}}\bng Aamra ekan iden Aaich\\ (what day are we in)\end{tabular} & DATE & TIME    & 0.86 & \begin{tabular}[c]{@{}c@{}}CURRENT\\LOCATION\end{tabular}& 0.11             \\ \hline
\end{tabular}
\end{table*}

\subsection{Performance Comparison}
To our knowledge, we are the first to research in the domain of intent classification tasks. Therefore, it is impossible to directly compare our work to other work of similar scope. To assess the viability of our work, we present a few earlier findings on the classification of English intent. It is clear from the preceding research in Table \ref{tab:comp} that our GAN-BnBERT model produced comparable results in detecting the intent of a user in Bengali.

\begin{table}[]
\centering
\caption{A comparison between our research and several past studies on intent classification}
\label{tab:comp}
\begin{tabular}{ccccc}
\hline
\multicolumn{1}{c}{\textbf{Reference Paper}} & \multicolumn{1}{c}{\textbf{{\begin{tabular}[c]{@{}c@{}}Proposed\\Model\end{tabular}}}} & \multicolumn{1}{c}{\textbf{Dataset}}                     & \multicolumn{1}{c}{\textbf{\begin{tabular}[c]{@{}c@{}}No. of  \\ Intent\end{tabular}}} & \multicolumn{1}{c}{\textbf{\begin{tabular}[c]{@{}c@{}}Accuracy\\ (\%)\end{tabular}}} \\ \hline
Larson et al. \cite{larson2019evaluation} & BERT                                         & \begin{tabular}[c]{@{}l@{}}CLINC150\\ (Full)\end{tabular} & 150    & 96.9 \\ \hline 
Mehri et al. \cite{mehri2019pretraining}& {\begin{tabular}[c]{@{}c@{}}CONVBERT\\+ MLM\end{tabular}}                               & \begin{tabular}[c]{@{}l@{}}CLINC150\\ (Full)\end{tabular} & 150    & 97.11\\ \hline
Chen et al. \cite{chen2019bert} & Joint BERT                                   & SNIPS \cite{coucke2018snips}                                                     & 7      & 98.6 \\ \hline 
Ours & {\begin{tabular}[c]{@{}c@{}}GAN-\\BnBERT\end{tabular}}                &BNIntent30 & 30    & 96.73\\ \hline
\end{tabular}
\end{table}

\section{Conclusion and Future work}
Categorizing intents is crucial for the success of a conversational agent. We present a fresh Bengali data set with 30 distinct classes for intent classification and a GAN-based architecture on top of BERT to develop a trustworthy intent categorization model for Bengali chatbots. To comprehend the black box behavior of the model, we also provide a full explanation of the proposed model and perform a detailed error analysis on our dataset. An extension of the dataset we propose, along with an out-of-scope intent categorization in Bengali, could be an exciting future direction for this research. Another potential further expansion is the inclusion of the slot-filling task will enhance the dataset comprehensively.

\bibliographystyle{IEEEtran}
\bibliography{references}

\end{document}